\documentclass[pdflatex,sn-nature]{sn-jnl}

\usepackage{graphicx}%
\usepackage{multirow}%
\usepackage{amsmath,amssymb,amsfonts}%
\usepackage{amsthm}%
\usepackage{mathrsfs}%
\usepackage[title]{appendix}%
\usepackage{xcolor}%
\usepackage{textcomp}%
\usepackage{manyfoot}%
\usepackage{booktabs}%
\usepackage{algorithm}%
\usepackage{algorithmicx}%
\usepackage{algpseudocode}%
\usepackage{listings}%
\usepackage{longtable}
\usepackage{array}
\usepackage{xcolor, colortbl}
\usepackage{booktabs}
\usepackage[inkscapelatex=false]{svg}
\usepackage{textcomp}
\usepackage{tabularx}
\usepackage{hyperref}
\usepackage{listings}
\usepackage{arydshln}
\definecolor{lightgray}{gray}{0.95}
\renewcommand{\arraystretch}{1.4}

\begin{document}

\title[Article Title]{Leveraging Open-Source Large Language Models for Clinical Information Extraction in Resource-Constrained Settings}

\author*[1]{\fnm{Luc} \sur{Builtjes}}\email{luc.builtjes@radboudumc.nl}

\author[1]{\fnm{Joeran} \sur{Bosma}}

\author[1]{\fnm{Mathias} \sur{Prokop}}

\author[1]{\fnm{Bram} \sur{van Ginneken}}

\author[1]{\fnm{Alessa} \sur{Hering}}

\affil[1]{\orgdiv{Diagnostic Image Analysis Group, Department of Medical Imaging}, \orgname{Radboud University Medical Center}, \orgaddress{\street{\city{Nijmegen}, \country{The Netherlands}}}}

\abstract{
Medical reports contain rich clinical information but are often unstructured and written in domain-specific language, posing challenges for information extraction. While proprietary large language models (LLMs) have shown promise in clinical natural language processing, their lack of transparency and data privacy concerns limit their utility in healthcare. This study therefore evaluates nine open-source generative LLMs on the DRAGON benchmark, which includes 28 clinical information extraction tasks in Dutch. We developed \texttt{llm\_extractinator}, a publicly available framework for information extraction using open-source generative LLMs, and used it to assess model performance in a zero-shot setting. Several 14 billion parameter models, Phi-4-14B, Qwen-2.5-14B, and DeepSeek-R1-14B, achieved competitive results, while the bigger Llama-3.3-70B model achieved slightly higher performance at greater computational cost. Translation to English prior to inference consistently degraded performance, highlighting the need of native-language processing. These findings demonstrate that open-source LLMs, when used with our framework, offer effective, scalable, and privacy-conscious solutions for clinical information extraction in low-resource settings.
}


\maketitle

\section{Introduction}\label{section:introduction}
Medical reports contain highly detailed patient information, including diagnoses, procedures, medications, and clinical observations, making them a valuable resource for data analysis for large-scale medical research \cite{dash2019big}. This density of clinically relevant information is especially valuable for developing artificial intelligence (AI) applications in healthcare, which depend on large, well-labeled datasets \cite{hosny2018artificial}. When processed effectively, these reports can yield a wide variety of training labels, supporting the development of accurate and generalizable AI models. 

The utility of medical reports is often limited by their unstructured textual format, which can vary significantly across institutions and individual practitioners \cite{meystre2008extracting}. Combined with the frequent use of domain-specific medical jargon, this lack of standardization presents a major challenge for information extraction, a critical step in converting raw clinical narratives into structured, machine-readable data.

Traditionally, the field of natural language processing (NLP) has relied on rule-based systems for information extraction, though these methods tend to struggle greatly with unstructured text \cite{adnan2019analytical}. The emergence of transformer-based models, such as BERT (Bidirectional Encoder Representations from Transformers) \cite{devlin2018bert}, enabled the extraction and structuring of meaningful data from more complex text. Domain-specific adaptations like Med-BERT \cite{rasmy2021med} have further refined these capabilities, achieving state-of-the-art performance in tasks like text classification and extraction. However, their effectiveness hinges on the availability of large quantities of labeled training data, which limits their scalability and adaptability for new tasks.

Recent advancements in generative Large Language Models (LLMs) have introduced a transformative shift in NLP. These models can be adapted to diverse tasks through the use of prompting techniques, reducing or even eliminating the reliance on task-specific training data. Their application in healthcare has already shown promise in areas such as clinical decision support \cite{ten2024chatgpt, kanjee2023accuracy, eriksen2023use, Fast2024}, medical text summarization \cite{van2024adapted}, and question answering \cite{singhal2023large, gilson2023does}.

However, a substantial portion of the current literature \cite{ten2024chatgpt, kanjee2023accuracy, eriksen2023use, singhal2023large, gilson2023does, shi2022language, li2024chatgpt, goh2025gpt, eriksen2024use, mitsuyama2025comparative, li2024comparing, park2024patient, noda2025gpt} is focused primarily on proprietary models such as OpenAI's GPT-4 \cite{achiam2023gpt}, which pose challenges related to transparency, reproducibility, and ethical concerns in clinical applications. These systems generally require transmitting data via an API to external servers where the models are hosted. This approach raises significant concerns under modern privacy regulations governing medical data, which mandate strict oversight over any information leaving hospital IT systems. Additionally, the training of many proprietary models on mostly undisclosed datasets raises ethical questions about data sourcing and contamination, privacy, and representativeness \cite{balloccu2024leakcheatrepeatdata}.

To address these limitations, the development and application of open-source LLMs have gained significant attention. These models offer researchers the opportunity to evaluate and adapt LLMs for medical tasks while ensuring greater accountability and control over input data by maintaining operations within local infrastructure. Open-source models also generally provide greater transparency regarding their pre-training datasets, enabling a clearer understanding of their limitations and biases \cite{kukreja2024survey}.

One of such limitations is the ability to effectively handle mid- to low-resource languages. Medical reports are predominantly written in the primary language of the care facility where they are produced. Proprietary models like GPT-4 benefit from extensive pre-training on datasets obtained through large-scale web scraping, which often include a variety of languages. In contrast, open-source LLMs are typically pre-trained on more curated datasets, leading to a disproportionate representation of high-resource languages such as English, Chinese, and Spanish. This imbalance results in a significant performance gap between these widely spoken languages and less common ones \cite{hasan2024large}. The challenge is further compounded when dealing with text rich in specialized jargon, such as the terminology found in medical contexts, where the disparity in linguistic resources becomes even more pronounced. Despite the practical significance of these issues, research on the performance of open-source models in such contexts remains limited \cite{vanroy2023language, gangavarapu2024introducing, wassie2024domain}.

The introduction of the Diagnostic Report Analysis: General Optimization of NLP (DRAGON) challenge \cite{bosma2025dragon} provides a valuable benchmark for addressing this issue. DRAGON includes 28,824 annotated medical reports from five care centers, covering 28 medically relevant information extraction tasks, such as classification, regression, and named entity recognition (NER), in the relatively uncommon Dutch language.

In this work, we present a systematic evaluation of several widely used open-source LLMs on domain-specific, resource-constrained language texts, with a focus on medical information extraction tasks. Our objective is to build a knowledge base identifying which models are most suitable for specific tasks in this setting, highlighting their strengths and limitations across various applications.

To support this evaluation, we developed a user-friendly and scalable framework that automates the application of open-source LLMs to diverse information extraction tasks on medical datasets in a language-agnostic manner. The framework enforces structured JavaScript Object Notation (JSON) output generation, enabling standardized and machine-readable outputs that facilitate both seamless evaluation and integration into downstream clinical or analytical pipelines. This design lowers the barrier to entry for deploying such models in complex, domain-specific contexts, while ensuring consistency and usability of the extracted information.

\begin{figure}[h]
    \centering
    \includegraphics[width=1\linewidth]{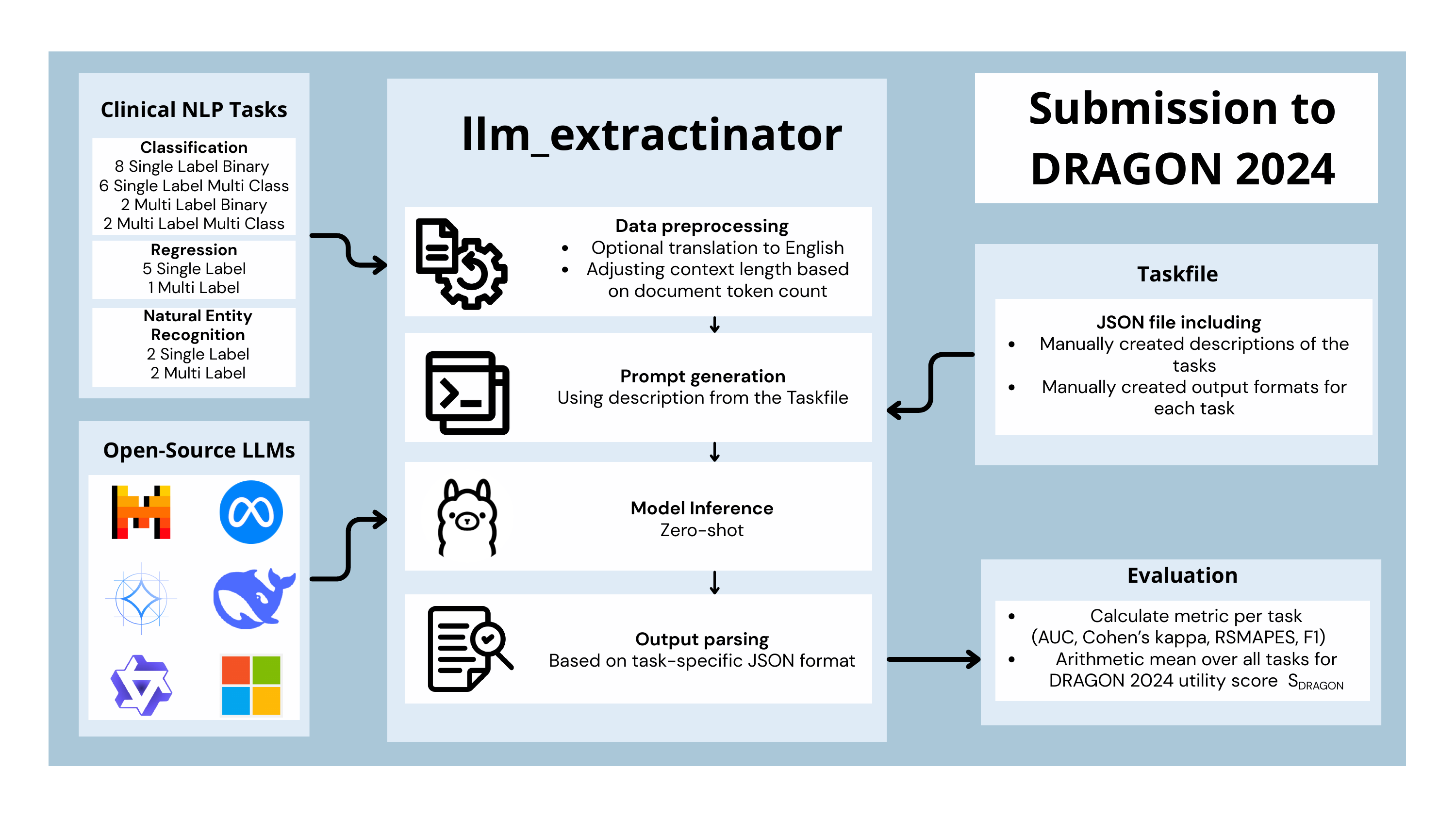}
    \caption{Overview of our submissions to the DRAGON 2024 challenge using the llm\_extractinator framework. We evaluate nine distinct Large Language Models (LLMs): Mistral-Nemo-12B(\url{https://mistral.ai/news/mistral-nemo/}), Llama-3.1-8B, Llama-3.2-3B, and Llama-3.3-70B \cite{dubey2024llama}, Gemma-2-2B and Gemma-2-9B \cite{team2024gemma}, Phi-4-14B \cite{abdin2024phi}, Qwen-2.5-14B \cite{yang2024qwen2}, and DeepSeek-R1-14B \cite{guo2025deepseek}. The input token length of each of the 28 clinical NLP tasks is measured, and model context windows are adapted accordingly. In three experiments, input text is translated to English using the LLM itself. For each task, we define a description and expected output format in a JSON-based Taskfile. This metadata guides prompt generation, which is followed by model inference and automatic output parsing. Task performance is assessed using the appropriate metric (AUC, Cohen’s kappa, RSMAPE, or F1), and the final DRAGON 2024 utility score $S_{\mathrm{DRAGON}}$ is computed as the arithmetic mean across all task metrics.}
    \label{fig:overview}
\end{figure}

The main contributions of our work are as follows:

\begin{enumerate} 
\item We introduce and publicly release \texttt{llm\_extractinator}, a scalable, language-agnostic, open-source framework for automating data extraction tasks with LLMs, designed for ease of use and broad applicability. It is available at \url{https://github.com/DIAGNijmegen/llm_extractinator}. 
\item We perform a comprehensive evaluation of nine widely used open-source LLMs on 28 medically relevant information extraction tasks using Dutch clinical reports in a zero-shot setting, as visually summarized in Figure~\ref{fig:overview}. This evaluation offers a realistic estimate for model performance and practical insights into the utility of generative LLMs in resource-constrained, domain-specific environments.
\end{enumerate}

By focusing on smaller, open-source generative models, our work contributes to bridging the gap between state-of-the-art AI capabilities and practical, real-world applications in healthcare. This study not only fills a critical void in the literature but also lays the foundation for future research in leveraging open-source LLMs for multilingual and resource-constrained medical environments.

\section{Results}\label{section:results}

We evaluated nine publicly available generative LLMs on the 28 tasks of the DRAGON challenge using the \texttt{llm\_extractinator} framework under zero-shot conditions. A representative input text for each tasks is available at: \url{https://github.com/DIAGNijmegen/dragon_sample_reports}. 

We followed the proposed metrics of the challenge organizers to quantify performance: area under the receiver-operating-characteristic curve (AUC) for binary classification, Cohen’s~$\kappa$ for multi-class classification, robust symmetric mean absolute percentage error score (RSMAPES) for regression, and $F_1$ score for NER. To facilitate model-level comparisons, we utilize the DRAGON 2024 utility score, $S_{\mathrm{DRAGON}}$, defined as the arithmetic mean of each model’s performance across all 28 tasks. The resulting score lies in the range [0,1], with 1 indicating perfect performance.

Additionally, the challenge organizers provide interpretability thresholds per metric which we use to categorize each result into one of six qualitative performance tiers: \emph{Excellent}, \emph{Good}, \emph{Moderate}, \emph{Poor}, \emph{Minimal}, or \emph{Fail}. Additional details on the metrics can be found in Supplementary Note 2. The full results are documented in Supplementary Note 3.

\subsection*{Model-Level Performance}

Model performance naturally clustered into three general tiers. The top-performing group consisted of Llama-3.3-70B ($S_{\mathrm{DRAGON}}=0.760$), Phi-4-14B (0.751), Qwen-2.5-14B (0.748), and DeepSeek-R1-14B (0.744). Llama-3.3-70B scores best with an \emph{Excellent} performance on 12 out of 28 tasks. Phi-4-14B achieved this performance on 10 out of 28 tasks, followed closely by Qwen-2.5-14B and DeepSeek-R1-14B, each with 9.

A second tier included Gemma2-9B and Mistral-Nemo-12B, both achieving $S_{\mathrm{DRAGON}}=0.688$, with \emph{Good} or better performance on roughly half the tasks. Llama-3.1-8B scored notably lower ($S_{\mathrm{DRAGON}}=0.588$), achieving \emph{Excellent} or \emph{Good} performance on just 7 tasks.

The lowest tier comprised Llama-3.2-3B ($S_{\mathrm{DRAGON}}=0.271$), which achieved only \emph{Minimal} to \emph{Fail} performance across all tasks. Gemma2-2B consistently failed to produce valid JSON outputs and thus could not be evaluated meaningfully.

\begin{figure}[h]
    \centering
    \includegraphics[width=1\linewidth]{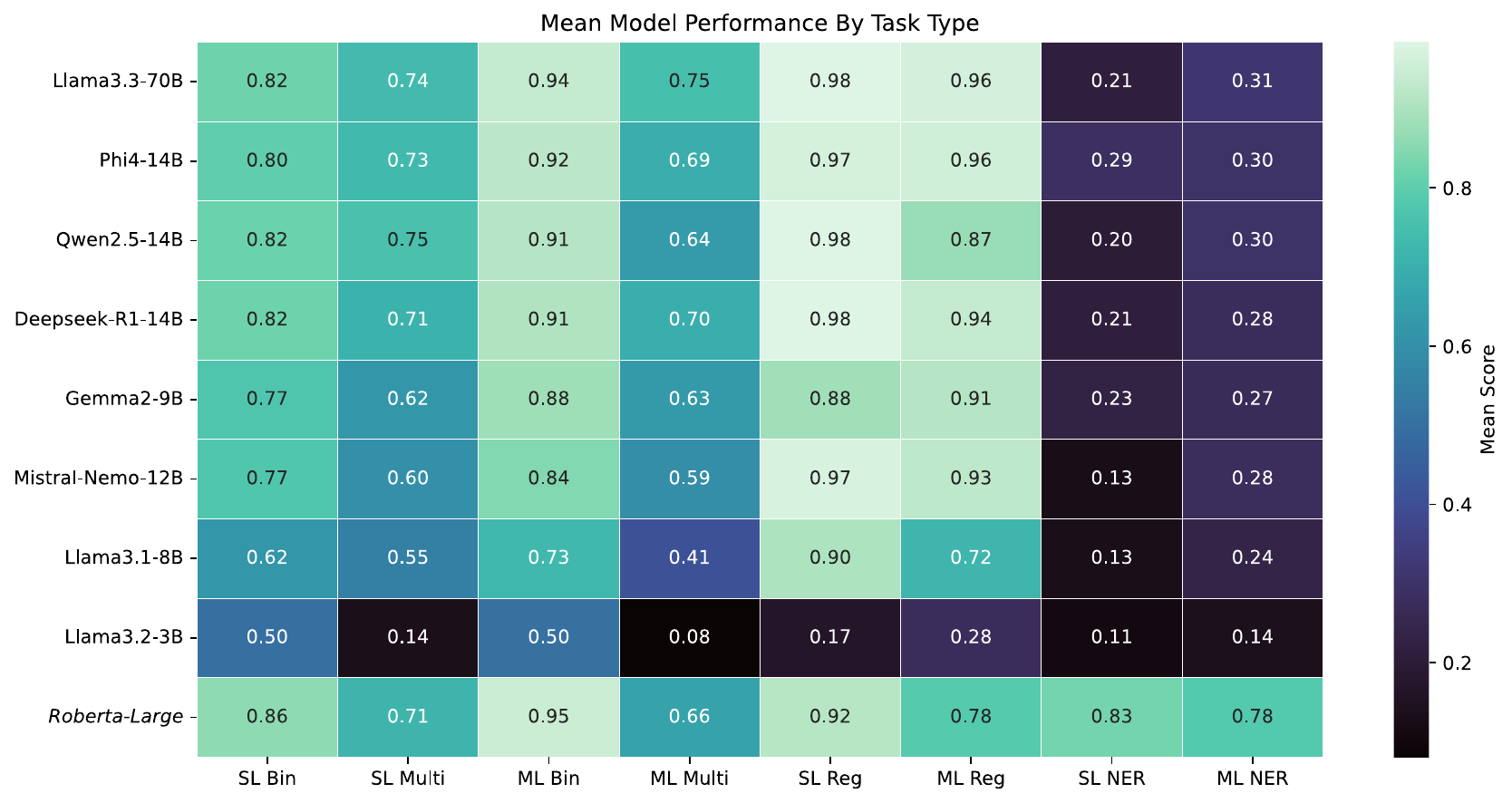}
    \caption{Heatmap illustrating the average performance of models across various task categories. Each cell shows the mean model score across tasks within a category. Scores range from 0 (worst) to 1 (best), except for multiclass classification tasks evaluated with Cohen’s kappa, which ranges from –1 (complete disagreement) to 1 (perfect agreement), with 0 indicating chance. For binary classification, 0.5 reflects chance-level performance. Task types are abbreviated as follows: \textbf{SL} = \textit{Single-label}, \textbf{ML} = \textit{Multi-label}, \textbf{Bin} = \textit{Binary classification}, \textbf{Multi} = \textit{Multi-class classification}, \textbf{Reg} = \textit{Regression}, \textbf{NER} = \textit{Named Entity Recognition}. The colormap represents average performance scores, with exact values annotated in each cell. The evaluation metric varies by task type: Area Under the Receiver Operating Characteristic Curve (AUC) is used for binary classification, Cohen’s Kappa for multi-class classification, Robust Symmetric Mean Absolute Percentage Error Score (RSMAPES) for regression, and F1 score for named entity recognition. The performance of the best performing baseline RoBERTa model of the challenge organizers (\url{https://grand-challenge.org/algorithms/dragon-roberta-large-domain-specific/}) is provided for reference.
}
    \label{fig:heatmap}
\end{figure}

Table~\ref{tab:model_summary} summarizes $S_{\mathrm{DRAGON}}$ scores alongside the number of tasks each model performed at each qualitative level. RoBERTa large with domain-specific pretraining, the best performing baseline model provided by the challenge organizers, is included for reference. Figure~\ref{fig:heatmap} visualizes average performance per task type across all models we tested. While the tiered structure is generally consistent across task types, certain models exhibit domain-specific strengths. For instance, Mistral-Nemo-12B performed comparably to top-tier models on regression tasks but underperformed on multi-label classification. Conversely, Gemma2-9B demonstrated relatively weaker performance on regression despite competitive results in other task types.

\begin{table}[ht]
\centering
\scriptsize
\renewcommand{\arraystretch}{1.2}
\setlength{\tabcolsep}{4pt}
\begin{tabular}{lccccccc}
\toprule
\textbf{Model} & $\mathbf{S_{\mathrm{DRAGON}}}$ & \textbf{Excellent} & \textbf{Good} & \textbf{Moderate} & \textbf{Poor} & \textbf{Minimal} & \textbf{Fail} \\
\midrule
LLaMA 3.3 70B         & 0.760 & 12 & 3 & 7 & 3 & 0 & 3 \\
Phi-4 14B             & 0.751 & 10 & 6 & 5 & 4 & 0 & 3 \\
Qwen 2.5 14B          & 0.748 &  9 & 7 & 6 & 2 & 1 & 3 \\
DeepSeek-R1 14B       & 0.744 &  9 & 6 & 5 & 5 & 1 & 2 \\
Gemma 2 9B            & 0.688 &  6 & 7 & 6 & 4 & 1 & 4 \\
Mistral-Nemo 12B      & 0.688 &  7 & 6 & 5 & 5 & 2 & 3 \\
LLaMA 3.1 8B          & 0.588 &  3 & 4 & 4 & 4 & 5 & 8 \\
LLaMA 3.2 3B          & 0.271 &  0 & 0 & 0 & 0 & 7 & 21 \\
\textit{RoBERTa Large} & 0.819 & 10 & 8 & 6 & 2 & 2 & 0 \\
\bottomrule
\end{tabular}
\caption{DRAGON 2024 utility scores ($S_{\mathrm{DRAGON}}$) and qualitative ratings across the 28 DRAGON tasks. RoBERTa Large is included as a reference, representing the current best-performing BERT-style baseline model provided by the challenge organizers (\url{https://grand-challenge.org/algorithms/dragon-roberta-large-domain-specific/}). Unlike our models, this baseline was trained directly on all 28 tasks and is not evaluated in a zero-shot setting.}
\label{tab:model_summary}
\end{table}

\subsection*{Task-Level Performance}

\begin{figure}[h]
    \centering
    \includegraphics[width=1\linewidth]{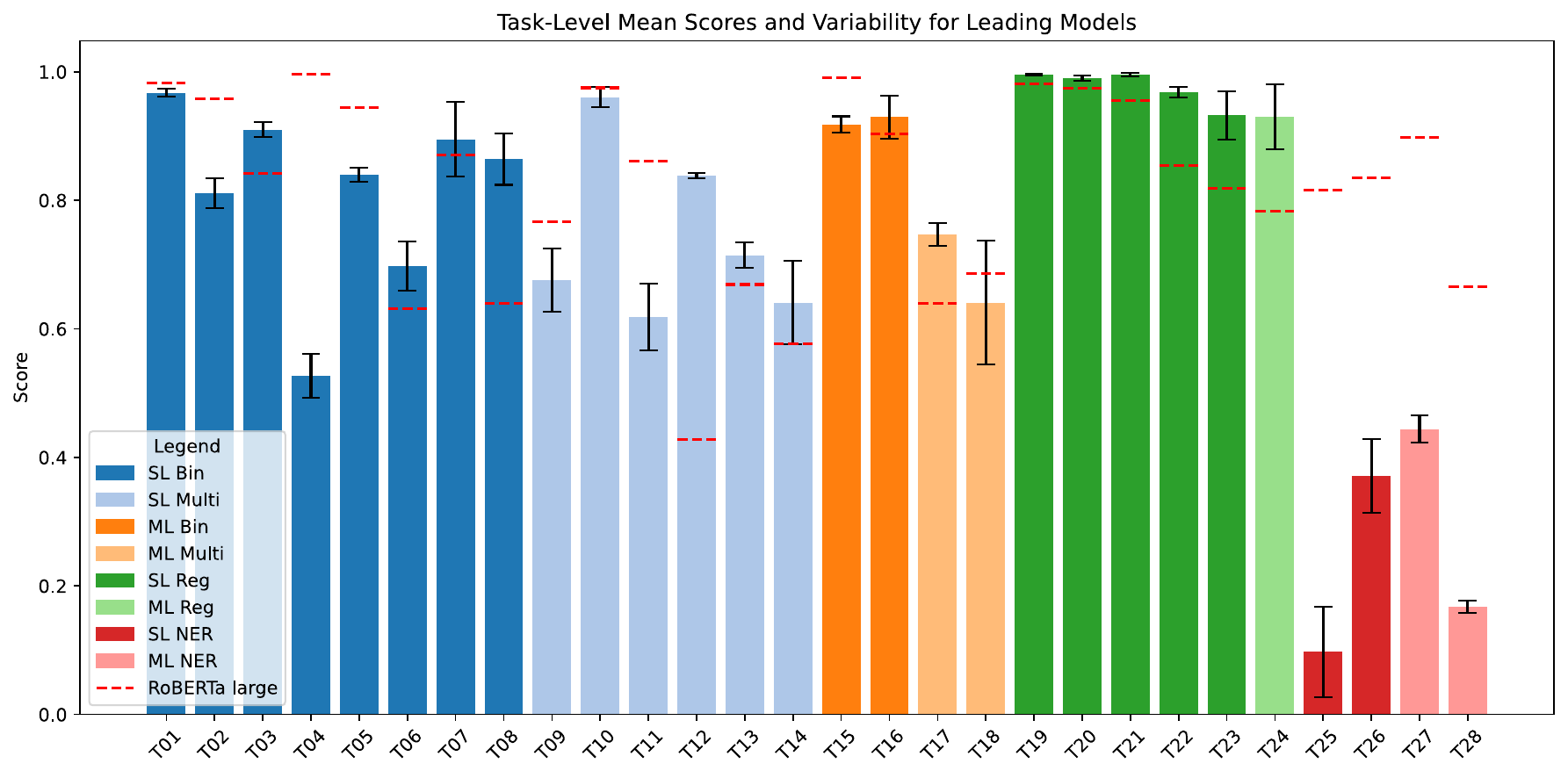}
    \caption{Mean performance scores and standard deviations for each of the 28 tasks, computed over the final evaluation metric across the top four performing models (Llama-3.3-70B, Phi-4-14B, Qwen-2.5-14B, and DeepSeek-R1-14B). Task types are color-coded and are abbreviated as follows: \textbf{SL} = \textit{Single-label}, \textbf{ML} = \textit{Multi-label}, \textbf{Bin} = \textit{Binary classification}, \textbf{Multi} = \textit{Multi-class classification}, \textbf{Reg} = \textit{Regression}, \textbf{NER} = \textit{Named Entity Recognition}. The mean performance per task of the best performing baseline RoBERTa model of the challenge organizers (\url{https://grand-challenge.org/algorithms/dragon-roberta-large-domain-specific/}) is provided as dotted red lines for reference.
}
    \label{fig:task_level_scores}
\end{figure}

Figure~\ref{fig:task_level_scores} shows task-specific performance distributions for the top four models. Across the six regression tasks, all models achieved scores $\geq 0.87$, with an average RSMAPES of $0.971$ and 22 out of 24 model–task combinations rated \emph{Excellent}. Binary classification tasks showed greater variability: while the group mean AUC of the four models over all tasks was 0.84, certain tasks (e.g., T04 and T06) saw performance near chance level for at least one model.

Ordinal classification tasks revealed broad score distributions, with Cohen’s~$\kappa$ values ranging from 0.51 to 0.98. The largest intra-task spread (T14, $\sigma=0.09$) illustrates the potential impact of model selection. Some tasks (e.g., T10 and T12) consistently produced high scores with low inter-model variability. In contrast, tasks T11, T14, and T18 displayed high variance and low scores, indicating task-level difficulty or sensitivity to model architecture.

NER performance was uniformly poor: none of the evaluated models exceeded an $F_1$ score of 0.47. The modal qualitative label for NER tasks was \emph{Fail}.

\subsection*{Comparison to BERT-style baseline model}

Table~\ref{tab:llama3_dragon_table} provides a detailed task-by-task comparison between the current top-performing model in the DRAGON 2024 challenge, DRAGON RoBERTa Large Domain-specific (\url{https://grand-challenge.org/algorithms/dragon-roberta-large-domain-specific/}), and our best performing model Llama-3.3 (\url{https://grand-challenge.org/algorithms/llm-extractinator-llama33/}). The better-performing model for each task is highlighted in bold. RoBERTa's results are reported as the mean and standard deviation from five-fold cross-validation, where the Llama-3.3 scores are derived from a single deterministic inference run with zero temperature.
`
\begin{table}[ht]
\centering
\scriptsize
\begin{tabular}{lcc}
\toprule
\textbf{Task} & \textbf{LLaMA3.3} & \textbf{RoBERTa Large} \\
\midrule
T01 & 0.971 & \textbf{0.983} ± 0.004 \\
T02 & 0.788 & \textbf{0.958} ± 0.008 \\
T03 & 0.923 & 0.842 ± 0.096 \\
T04 & 0.500 & \textbf{0.996} ± 0.001 \\
T05 & 0.840 & \textbf{0.944} ± 0.010 \\
T06 & \textbf{0.708} & 0.631 ± 0.042 \\
T07 & \textbf{0.955} & 0.870 ± 0.040 \\
T08 & \textbf{0.883} & 0.640 ± 0.050 \\
T09 & 0.619 & \textbf{0.767} ± 0.039 \\
T10 & 0.978 & 0.975 ± 0.003 \\
T11 & 0.669 & \textbf{0.861} ± 0.003 \\
T12 & \textbf{0.842} & 0.428 ± 0.057 \\
T13 & \textbf{0.736} & 0.669 ± 0.022 \\
T14 & 0.573 & 0.577 ± 0.009 \\
T15 & 0.917 & \textbf{0.991} ± 0.010 \\
T16 & \textbf{0.959} & 0.903 ± 0.015 \\
T17 & \textbf{0.767} & 0.639 ± 0.074 \\
T18 & \textbf{0.732} & 0.686 ± 0.015 \\
T19 & \textbf{0.995} & 0.981 ± 0.002 \\
T20 & \textbf{0.991} & 0.974 ± 0.002 \\
T21 & \textbf{0.993} & 0.955 ± 0.012 \\
T22 & \textbf{0.976} & 0.854 ± 0.003 \\
T23 & \textbf{0.955} & 0.818 ± 0.012 \\
T24 & \textbf{0.961} & 0.783 ± 0.003 \\
T25 & 0.028 & \textbf{0.816} ± 0.007 \\
T26 & 0.401 & \textbf{0.835} ± 0.003 \\
T27 & 0.467 & \textbf{0.898} ± 0.010 \\
T28 & 0.161 & \textbf{0.666} ± 0.035 \\
\midrule
\textbf{Score} & 0.760 & \textbf{0.819} ± 0.021 \\
\bottomrule
\end{tabular}
\caption{Task-wise comparison between LLaMA3.3-70B (\url{https://grand-challenge.org/algorithms/llm-extractinator-llama33/}) and DRAGON RoBERTa Large Domain-specific (\url{https://grand-challenge.org/algorithms/dragon-roberta-large-domain-specific/}). RoBERTa scores include standard deviations based on 5-fold cross-validation. A score is bolded when it is higher than the other model's and lies outside of RoBERTa's standard deviation.}
\label{tab:llama3_dragon_table}
\end{table}

The RoBERTa model achieved a higher overall DRAGON 2024 utility score ($S_{\mathrm{DRAGON}} = 0.819 \pm 0.021$) than any of the generative LLMs. However, across the 28 tasks, DRAGON RoBERTa Large Domain-specific achieved higher scores than Llama-3.3 on only 11 tasks (T01, T02, T04, T05, T09, T11, T15, T25, T26, T27, T28), with performance differences exceeding the upper bound of RoBERTa’s standard deviation, whereas the Llama model scored higher than RoBERTa on 14 tasks (T06, T07, T08, T12, T13, T16, T17, T18, T19, T20, T21, T22, T23, T24). For the remaining three tasks (T03, T10, and T14), the score of the generative LLM fell within one standard deviation of RoBERTa's mean score. 

\subsection*{Effect of In-Context Translation}

To assess the impact of in-context translation, we compared model performance on Dutch inputs with and without prior translation to English by the LLMs themselves. This translation strategy led to statistically significant performance degradation across all tested models. For Mistral-Nemo-12B, the $S_{\mathrm{DRAGON}}$ score dropped from 0.688 to 0.573 ($\Delta=-0.11$, $p<0.001$), for Phi-4-12B from 0.751 to 0.533 ($\Delta=-0.22$, $p<0.001$), and for Llama-3.1-8B, from 0.588 to 0.337 ($\Delta=-0.25$, $p<0.001$). These results indicate that in-context translation consistently harms downstream performance.

\begin{figure}[ht]
    \centering
    \includegraphics[width=1\linewidth]{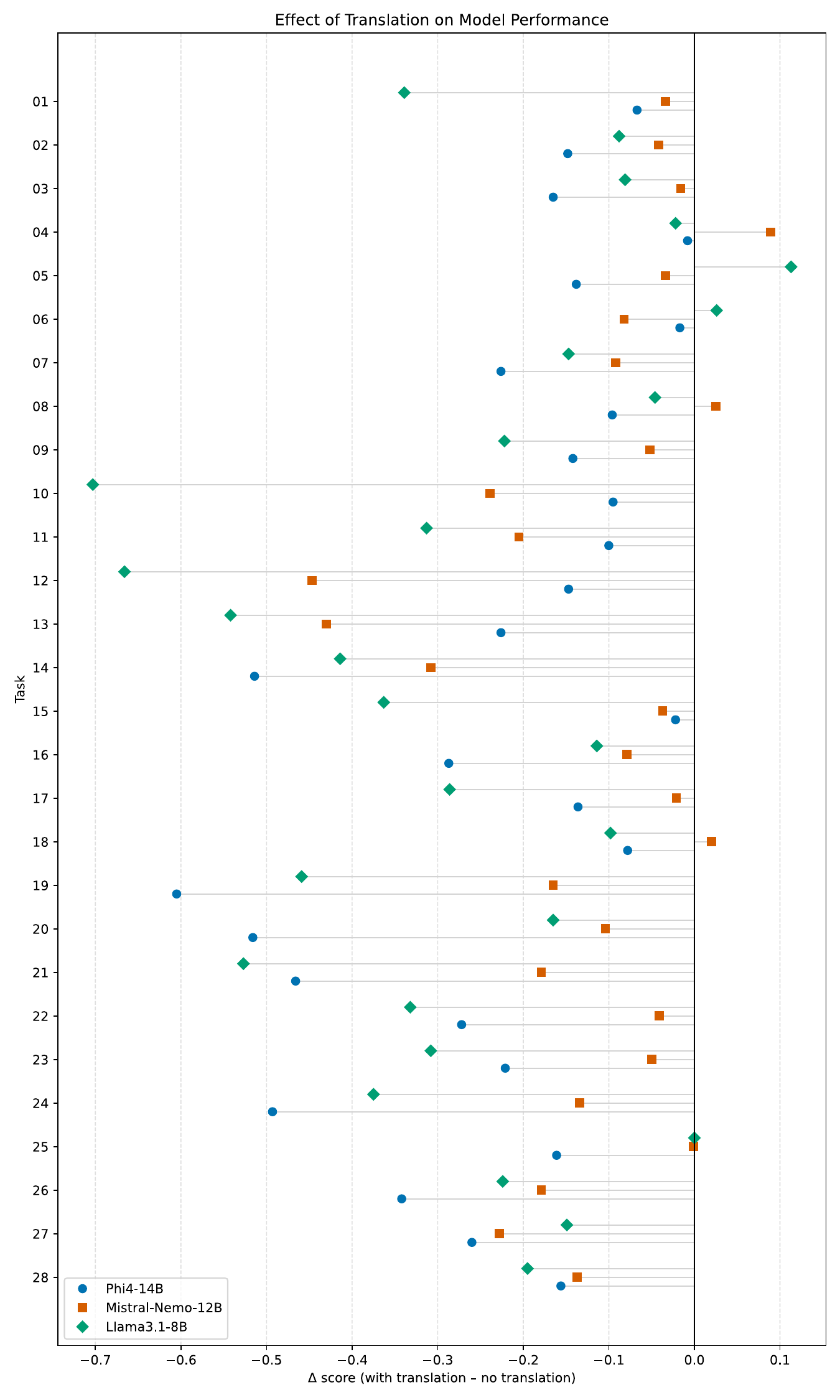}
    \caption{Impact of Machine Translation on LLM Performance Across 28 Clinical NLP Tasks in the DRAGON Challenge. This plot illustrates the performance deltas (with translation - without translation) for Phi-4-14B, Mistral-NeMo-12B, and Llama-3.1-9B. Negative deltas indicate that translation on average degrades model performance across tasks.}
    \label{fig:translation_results}
\end{figure}

Figure~\ref{fig:translation_results} details the differences in performance for all models where translation was tested, showing consistent task-level reductions in the relevant performance metric. These results suggest that translation-induced noise undermines clinical information extraction accuracy, underscoring the importance of native-language inference for domain-specific tasks.

\section{Discussion}\label{section:discussion}

In this study, we evaluated the zero-shot performance of nine widely used open-source LLMs using the llm\_extractinator framework on the DRAGON challenge, a Dutch clinical NLP benchmark. Our results highlight both the promise and limitations of deploying such models for real-world information extraction tasks in healthcare.

We found that models with around 14B parameters, including Phi-4, Qwen-2.5, and DeepSeek-R1, performed well across most tasks, achieving average DRAGON utility scores near 0.75. The Llama-3.3-70B model outperformed all others with a utility score of 0.76, consistent with prior findings that larger models tend to generalize better \cite{kaplan2020scaling, ahuja2023megaverse}. However, this improvement came with significant computational cost and only translated into higher task-level performance in 11 of 28 cases. This suggests that performance gains from scaling are not uniform across task types, and that deploying larger models is most justifiable when computational resources are readily available and the marginal performance gains are considered worthwhile.

Regression tasks, such as extracting lesion sizes or PSA levels, were a relative strength for all tested LLMs. These results contrast with the weaker performance by fine-tuned BERT-style models on the multi-label regression tasks in particular. Generative models appear to handle numeric value reproduction especially well due to their copy-and-reason capabilities. This aligns with prior intuitions that generative models retain quantitative tokens during inference whereas this is more difficult for encoder-based models \cite{wallace2019nlp}.

Performance declined markedly on classification tasks and collapsed on NER. Even the strongest models achieved F1 scores below 0.5 on the latter. This underperformance was likely exacerbated by the token-level output format required by the DRAGON challenge. Generative models are not naturally suited for generating sparsely populated token-level lists, and our structured prompting followed by post-processing likely introduced conversion errors. It has been shown in other work that more suitable evaluation formats can yield good performance on similar tasks \cite{wiest2024anonymizing}. As such, our results represent only a conservative estimate of model capabilities on these tasks.

Additionally, certain tasks were inherently unsuited to zero-shot evaluation and were therefore unlikely to succeed. For instance, Task 04 (Skin histopathology case selection) asked models to determine whether a pathology report should be excluded based on vague criteria such as being incomplete or lacking a definitive diagnosis. In the intended use case, where models are trained on labeled examples, such patterns could be learned. However, in a zero-shot setting, where no task-specific feedback or examples are available, the prompt alone provides insufficient guidance. This limitation is reflected in the near-random performance of even the top-performing models on this task.

Annotation quality also influenced model performance. As reported by the DRAGON organizers, some tasks showed relatively low inter-annotator agreement. Notably, Task 06 (histopathology cancer origin) had a Krippendorff’s Alpha of 0.333, Task 14 (textual entailment) scored 0.550, Task 17 (PDAC attributes) 0.677, and Task 18 (hip osteoarthritis scoring) 0.557. Inconsistent labeling likely contributed to higher variance across models and limited overall accuracy, even for top-performing systems.

While the DRAGON leaderboard is currently led by fine-tuned encoder models, our zero-shot evaluation of generative models paints a more nuanced picture. Although Llama-3.3 trailed the top-performing RoBERTa-based model overall (0.760 vs 0.819 utility score), this difference is primarily due to strong relative performance of the RoBERTa model on NER tasks and Task 04. Excluding these tasks shifts the average score in favor of Llama-3.3. Its $S_{\mathrm{DRAGON}}$ rises to 0.858, while RoBERTa’s drops to 0.814. This suggests complementary strengths: encoder models excel at token-level classification, while generative LLMs are better suited to structured inference and regression tasks.

Importantly, these comparisons must be contextualized within the operational and data constraints of real-world deployments. Fine-tuned RoBERTa models require supervised training on labeled data for each task, and are tightly coupled to their respective training distributions. By contrast, Llama-3.3 and other generative LLMs were evaluated strictly in a zero-shot setting, without any parameter updates or task-specific examples in-context. That they perform comparably under these conditions, sometimes even exceeding RoBERTa's performance, suggests that generative models are becoming increasingly viable alternatives for scalable, plug-and-play clinical NLP, especially in settings where labeled data is scarce or task requirements evolve frequently.

Our experiments also reveal that translating Dutch clinical text into English before inference led to reduced performance, despite theoretical advantages from English-centric training corpora. This supports growing evidence that translation introduces artifacts and dilutes clinical nuance. The findings argue against translation-based workarounds and reinforce the importance of native language support in multilingual clinical NLP.

Another key insight from our analysis is the significant negative impact of translating Dutch medical texts into English prior to inference. Across the board, naïve translation consistently degraded performance. While the literature presents mixed findings depending on the context, with some studies reporting benefits from pre-translation \cite{mondshine2025beyond, chen2023translation} while others observe better outcomes without it \cite{intrator2024breaking}, our results suggest that translation on this dataset introduces artifacts and erodes clinical nuance. These findings underscore the need for robust native-language support in clinical NLP tools.

Smaller models, such as Llama-3.2-3B and Gemma-2-2B, consistently failed across tasks, producing nonsensical outputs. This establishes a practical lower bound for model scale for zero-shot clinical NLP in a non-English language. Our \texttt{llm\_extractinator} framework supports efficient inference of larger models on consumer-grade GPUs. This effectively eliminates the need to rely on underpowered models. While smaller models may offer marginal improvements in inference speed, the trade-off in output quality is steep: the risk of generating entirely unusable results outweighs any computational performance gains.

This study has several limitations. First, a uniform prompting approach was used across tasks and models, without extensive task-specific engineering. While this supports reproducibility, it likely underestimates achievable performance. Second, our evaluation was limited to zero-shot settings in Dutch. Generalizability to other languages remains to be tested and there is room for future research to explore the effects of few-shot prompting \cite{brown2020language}, lightweight instruction-tuning \cite{wei2021finetuned}, or retrieval-augmented generation \cite{lewis2020retrieval} on model performance. Finally, due to resource constraints, we only evaluated one model over 15B parameters and did not include any of the largest open-source LLMs. Future work should explore their capabilities, especially in high-resource settings.

In summary, this work demonstrates that open-source generative LLMs can serve as powerful tools for medical information extraction in Dutch. With minimal infrastructure or labeled data, several models approach or surpass fine-tuned encoder baselines in clinical NLP tasks. By streamlining this process through our llm\_extractinator framework, we lower the barrier to applying these models in real-world clinical research.
\section{Methods}\label{section:methods}

\subsection*{llm\_extractinator}\label{section:framework}
To enable efficient information extraction from unstructured text using generative LLMs, we developed the \textbf{\texttt{llm\_extractinator}} framework, available on GitHub (\url{https://github.com/DIAGNijmegen/llm_extractinator}) and as a pip-installable Python package.

To maximize ease-of-use, the framework requires only two user-provided inputs: input data and a Taskfile. This Taskfile, formatted in JSON, specifies both the task description and the desired output structure. The output structure defines the target JSON format for the extracted information. To further streamline the design process, a user-friendly web application is included with the package, allowing users to interactively design and preview the desired output format without any manual code writing.

The backbone of \texttt{llm\_extractinator} utilizes Ollama (\url{https://ollama.com}). This is an open-source model hub for LLMs that facilitates model weight retrieval, local server hosting for inference, and automatic distribution of computational load across available resources. Our framework is plug-and-play compatible with any LLM available on the Ollama model hub or any customized model otherwise created using an Ollama Modelfile.

To optimize computational efficiency, \texttt{llm\_extractinator} dynamically adjusts the model context length based on dataset characteristics. This setting refers to the maximum number of tokens that the model can process simultaneously in a single inference step. Setting the context length too high can lead to unnecessary memory usage and slower inference times, while setting it too low may result in incomplete processing of the input data. Two operational modes are available: when the setting is specified as \texttt{max}, the maximum token length observed within the dataset is used. Alternatively, when set to \texttt{split}, the dataset is partitioned into subsets according to a specified fraction, with each subset assigned an appropriate context length. This approach allows the bulk of shorter texts to be processed efficiently, while longer outlier reports are accommodated without loss of data fidelity.

Prompt construction in \texttt{llm\_extractinator} is performed through LangChain (\url{https://www.langchain.com/}), which is used to combine the input data and task instructions from the Taskfile into a formatted prompt. During this stage, zero-shot chain-of-thought prompting \cite{wei2022chain} is applied, a technique wherein the LLM is encouraged to explicitly articulate intermediate reasoning steps prior to delivering a final answer. An optional translation module enables the automatic translation of input text into English by the LLM prior to task execution.

Following model inference, the output is validated against the user-specified output schema. Outputs that do not conform are automatically resubmitted for reformatting, with up to three attempts permitted by default. If after three attempts the output remains non-compliant, a placeholder entry (either random or empty) is generated to ensure the continuity of automated evaluation workflows. These instances are subsequently flagged for manual review.

\subsection*{DRAGON challenge}\label{section:dragon}
To evaluate model performance, we applied our framework to the DRAGON challenge, a benchmark initiative for clinical NLP tasks in Dutch (\url{https://dragon.grand-challenge.org/}). The DRAGON dataset comprises 28,824 annotated medical reports collected from five Dutch healthcare institutions, covering 28 clinically relevant tasks. These tasks span a diverse range of categories: 8 single-label binary classification tasks, 6 single-label multi-class classification tasks, 2 multi-label binary classification tasks, 2 multi-label multi-class classification tasks, 5 single-label regression tasks, 1 multi-label regression task, 2 single-label NER tasks, and 2 multi-label NER tasks. A complete overview of the tasks is provided in Figure \ref{fig:DRAGON_overview} and in Supplementary Note 1. A representative input text for each tasks is available at: \url{https://github.com/DIAGNijmegen/dragon_sample_reports}. 

\begin{figure}[ht]
    \centering
    \includegraphics[width=1\linewidth]{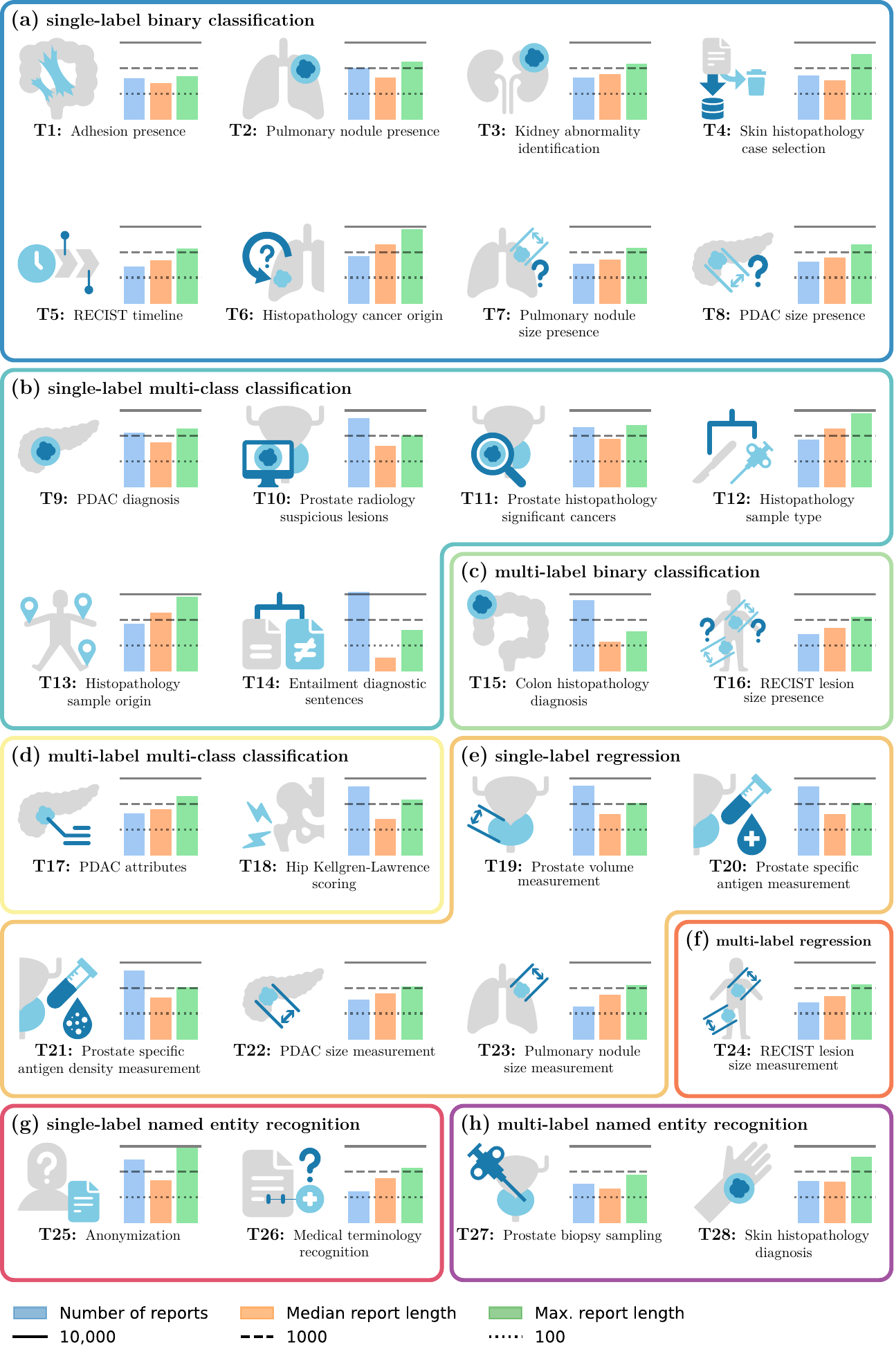}
    \caption{An overview of the 28 different tasks of the DRAGON challenge grouped by task type. The bar graphs show the number of reports, the median report length, and the maximum report length in each dataset based on tokenization using an xlm-roberta-base tokenizer. Figure reproduced with permission from Bosma et al. (2025) \cite{bosma2025dragon}.}
    \label{fig:DRAGON_overview}
\end{figure}

The challenge is hosted on the Grand Challenge platform \cite{meakin2024grand}, a fully cloud-based environment powered by Amazon Web Services. The full challenge workflow involves two stages: fine-tuning BERT-like models on a provided training and validation dataset, followed by inference on a separate test set. However, since our study aims to evaluate the zero-shot capabilities of generative LLMs, we bypassed the fine-tuning phase and conducted direct inference via prompting.

The outputs generated by the models were post-processed into the JSON format required for automatic evaluation. These predictions were then evaluated against the ground truth test labels, producing task-specific performance scores. For the binary classification tasks (T1–T8), performance was assessed using the Area Under the Receiver Operating Characteristic Curve (AUC). For the multi-class classification tasks (T9–T14), performance was measured using either unweighted or linearly weighted Cohen’s Kappa, depending on the task. Multi-label classification tasks (T15–T18) were evaluated using either macro-averaged AUC or unweighted Kappa. Regression tasks (T19–T24) used the Robust Symmetric Mean Absolute Percentage Error Score (RSMAPES) with task-specific tolerance margins. Named entity recognition tasks (T25–T28) were evaluated using macro or weighted F1 scores.

The individual task scores were aggregated into the DRAGON 2024 utility score $S_{\mathrm{DRAGON}}$, computed as the arithmetic mean of the performance metrics across all 28 tasks. While the standard DRAGON evaluation protocol recommends five test runs using different random seeds to account for sampling variability, we opted for a single test run by setting the model's sampling temperature to 0. This configuration enforces fully deterministic outputs in token generation, allowing for more stable, direct comparisons of zero-shot model performance. This approach also offers practical advantages by substantially reducing computational costs, as it eliminates the need for four additional inference runs per model.

\subsection*{Models}\label{section:models}
We evaluated nine widely used open-source multilingual LLMs available through the Ollama model hub: Llama3.1-8B, Llama3.2-3B, and Llama3.3-70B \cite{dubey2024llama}, Gemma2-2B and Gemma2-9B \cite{team2024gemma}, Phi4-14B \cite{abdin2024phi}, Qwen2.5-14B\cite{yang2024qwen2}, DeepSeek-R1-14B\cite{guo2025deepseek}, and Mistral-NeMo(\url{https://mistral.ai/news/mistral-nemo/}). For inference efficiency, all models were run in 4-bit quantized format. Specifically, we used the q4\_0 quantization scheme for Mistral-Nemo and Gemma2, and q4\_K\_M for the remaining models. These configurations reflect the default quantization settings provided by the Ollama model hub.

Our primary focus was on models with fewer than 15 billion parameters to ensure practical feasibility within typical hospital IT environments. All such models can be run on consumer-grade GPUs with 12GB of VRAM when quantized to 4-bit precision. Although the llm\_extractinator framework supports CPU offloading to accommodate larger models on limited hardware, this results in significant reductions in processing speed, rendering it impractical for clinical deployment. Given that most healthcare facilities lack ready access to high-performance GPUs, we prioritized smaller models for this study. Nevertheless, by also evaluating one larger model, we aimed to provide an informed benchmark for institutions with more substantial computational resources.

\subsection*{Prompting strategies}\label{section:prompting}
All experiments were conducted under a strict zero-shot setting, meaning that no task-specific fine-tuning was employed and no examples were provided in-context. This approach tests the models' inherent capabilities to perform unfamiliar medical tasks based solely on their pretrained knowledge. To encourage model reasoning, we consistently applied zero-shot chain-of-thought prompting across all tasks. A full list of all prompts used is provided in Supplementary Note 4.

Moreover, we investigated the effect of translating the original Dutch reports into English prior to inference, given that the LLMs were predominantly trained on English corpora. This intermediate translation step was hypothesized to improve performance by aligning the input language with the models' primary training distribution, thus potentially reducing comprehension errors arising from linguistic mismatch.

\subsection*{Statistical analysis}
To evaluate the effect of in-context translation on model performance, we conducted pairwise comparisons between performance scores obtained with and without translation across identical input sets. For each language model, we computed task-level scores under both conditions and assessed statistical significance using paired statistical tests.

Normality of score differences was evaluated using the Shapiro–Wilk test. If the differences were normally distributed ($p>0.05$), a paired t-test was applied; otherwise, the non-parametric Wilcoxon signed-rank test was used. All tests were two-tailed with a significance threshold of $p<0.05$. Mean scores and absolute performance differences ($\Delta$) were reported alongside p-values for each model. Statistical analyses were performed using Python 3.11 with SciPy v1.15.3.

\clearpage
\bibliography{sn-bibliography}

%\clearpage
%\section*{Figures}
%\input{figures/overview/overview}
%\clearpage
%\input{figures/heatmap/heatmap}
%\clearpage
%\input{figures/task_level_scores/task_level_scores}
%\clearpage
%\input{figures/translation_results/translation_results}
%\clearpage
%\input{figures/DRAGON_overview/DRAGON_overview}

%\clearpage
%\section*{Tables}
%\input{tables/model_summary}
%\clearpage
%\input{tables/dragon_compare}
\clearpage
\section*{Supplementary Note 1: Overview of DRAGON Tasks}
\label{supp1}
The DRAGON benchmark \cite{bosma2025dragon} comprises 28 diverse clinical tasks across modalities such as radiology, pathology, and structured reports. Tasks are categorized by learning setup—single-label (SL) vs. multi-label (ML), classification (binary or multi-class), regression, and named entity recognition (NER)—and evaluated using domain-appropriate metrics (e.g., AUC, Kappa, F1, RSMAPES). This table summarizes all tasks included in the benchmark.

\begin{table}[htbp]
\centering
\caption{Overview of DRAGON tasks. \textbf{SL} = \textit{single-label}, \textbf{ML} = \textit{multi-label}, \textbf{Bin} = \textit{binary}, \textbf{MC} = \textit{multi-class}, \textbf{CLF} = \textit{classification}, \textbf{Reg} = \textit{regression}, \textbf{NER} = \textit{named entity recognition}, \textbf{AUC} = \textit{Area Under the Receiver Operator Characteristic Curve}, \textbf{RSMAPES} = \textit{Robust Symmetric Mean Absolute Percentage Error Score}, \textbf{PDAC} = \textit{Pancreatic Ductal Adenocarcinoma}, \textbf{RECIST} = \textit{Response Evaluation Criteria In Solid Tumors}.}
\renewcommand{\arraystretch}{1.2}
\begin{tabularx}{\textwidth}{|c|X|c|c|}
\hline
\rowcolor[HTML]{EFEFEF} \textbf{ID} & \textbf{Task Name} & \textbf{Task Type} & \textbf{Evaluation Metric} \\ \hline
T1 & Adhesion presence & SL Bin CLF & AUC \\ \hline
T2 & Pulmonary nodule presence & SL Bin CLF & AUC \\ \hline
T3 & Kidney abnormality identification & SL Bin CLF & AUC \\ \hline
T4 & Skin histopathology case selection & SL Bin CLF & AUC \\ \hline
T5 & RECIST timeline & SL Bin CLF & AUC \\ \hline
T6 & Histopathology cancer origin & SL Bin CLF & AUC \\ \hline
T7 & Pulmonary nodule size presence & SL Bin CLF & AUC \\ \hline
T8 & PDAC size presence & SL Bin CLF & AUC \\ \hline
T9 & PDAC diagnosis & SL MC CLF & Unweighted Kappa \\ \hline
T10 & Prostate radiology suspicious lesions & SL MC CLF & Linearly Weighted Kappa \\ \hline
T11 & Prostate histopathology significant cancers & SL MC CLF & Linearly Weighted Kappa \\ \hline
T12 & Histopathology tissue type & SL MC CLF & Unweighted Kappa \\ \hline
T13 & Histopathology tissue origin & SL MC CLF & Unweighted Kappa \\ \hline
T14 & Entailment of diagnostic sentences & SL MC CLF & Linearly Weighted Kappa \\ \hline
T15 & Colon histopathology diagnosis & ML Bin CLF & Macro AUC \\ \hline
T16 & RECIST lesion size presence & ML Bin CLF & AUC \\ \hline
T17 & PDAC attributes & ML MC CLF & Unweighted Kappa \\ \hline
T18 & Hip Kellgren–Lawrence scoring & ML MC CLF & Unweighted Kappa \\ \hline
T19 & Prostate volume extraction & SL Reg & RSMAPES ($\varepsilon$ = 4 cm$^3$) \\ \hline
T20 & PSA extraction & SL Reg & RSMAPES ($\varepsilon$ = 0.4 ng/mL) \\ \hline
T21 & PSA density extraction & SL Reg & RSMAPES ($\varepsilon$ = 0.04 ng/mL$^2$) \\ \hline
T22 & PDAC size measurement & SL Reg & RSMAPES ($\varepsilon$ = 4 mm) \\ \hline
T23 & Pulmonary nodule size measurement & SL Reg & RSMAPES ($\varepsilon$ = 4 mm) \\ \hline
T24 & RECIST lesion size measurement & ML Reg & RSMAPES ($\varepsilon$ = 4 mm) \\ \hline
T25 & Anonymization & SL NER & Macro F1 \\ \hline
T26 & Medical terminology recognition & SL NER & F1 \\ \hline
T27 & Prostate biopsy sampling & ML NER & Weighted F1 \\ \hline
T28 & Skin histopathology diagnosis & ML NER & Weighted F1 \\ \hline
\end{tabularx}
\end{table}

\clearpage

\section*{Supplementary Note 2: Metrics \& Result Interpretation}
\label{supp2}
The metrics used to determine performance per task follow those used in the DRAGON challenge \cite{bosma2025dragon}. The class-neutral area under the receiver operating characteristic curve (AUC) is used for binary classification tasks. For multi-labeled cases, results are either pooled before calculating a single AUC value (Task16), or AUC is calculated per label and then averaged as Macro AUC (Task15). For ordinal multi-class, either linearly weighted or unweighted Cohen's kappa are used, depending on whether the classes are ordinal or non-ordinal respectively. For named entity recognition tasks we employed variations of macro F1. Macro F1 was used for the single-label tasks, weighted F1 for the multi-label ones. Regression tasks used the Robust Symmetric Mean Absolute Percentage Error Score (RSMAPES), defined as

\begin{equation*}
    1 - \sum_{i}^{N} \frac{|y-\hat{y}|}{|\hat{y}| + |y| + \epsilon}
\end{equation*}

where $N$ is the number of cases, $y$ is the target value, $\hat{y}$ is the predicted value, and $\epsilon$ is a positive value. The value of $\epsilon$ was set depending on the task.

To standardize the interpretation of model performance, qualitative thresholds were applied to each evaluation metric. These thresholds align with those used in the DRAGON Challenge paper, providing a comparable framework for interpretation. The table below summarizes the value ranges corresponding to qualitative labels across all metrics used in this study.

\begin{table}[h]
\centering
\begin{tabular}{|c|c|}
\hline
\textbf{Range} & \textbf{Performance Level} \\
\hline
\multicolumn{2}{|c|}{\textbf{Kappa, F1, RSMAPES}} \\
\hline
0.90 -- 1.00  & Excellent \\
0.80 -- 0.90  & Good      \\
0.60 -- 0.80  & Moderate  \\
0.40 -- 0.60  & Poor      \\
0.21 -- 0.40  & Minimal   \\
$<$ 0.21      & Fail      \\
\hline
\multicolumn{2}{|c|}{\textbf{AUC}} \\
\hline
0.90 -- 1.00  & Excellent \\
0.80 -- 0.90  & Good      \\
0.70 -- 0.80  & Moderate  \\
0.65 -- 0.70  & Poor      \\
0.60 -- 0.65  & Minimal   \\
$<$ 0.60      & Fail      \\
\hline
\end{tabular}
\caption{Interpretation thresholds for model evaluation metrics.}
\label{tab:metric-interpretation}
\end{table}

\clearpage

\section*{Supplementary Note 3: Complete results}
\label{supp3}
Table \ref{tab:model_mt_comparison} reports model performance across all 28 tasks included in DRAGON 2024 \cite{bosma2025dragon}. Scores are shown for Phi4, Qwen2.5, Deepseek, Gemma2, Mistral, Llama3.1, and Llama3.2. The best score for each task is bolded. The DRAGON utility score ($S_{\mathrm{DRAGON}}$), defined as the arithmetic mean performance across all tasks, is listed in the final row.

Table \ref{tab:model_performance_translation_abbrev} shows results for the same 28 tasks when processed via machine translation to English, evaluated on Phi4, Mistral and Llama3.1, compared to the result without it. The highest score per task is bolded, and $S_{\mathrm{DRAGON}}$ is included for comparison.

Table \ref{tab:model_legend} lists the full model names corresponding to the abbreviations used in the results tables.

\begin{table}[h]
\centering
\caption{Performance across 28 benchmark tasks. For each task, the best score is bolded. The DRAGON 2024 utility score ($S_{\mathrm{DRAGON}}$) is calculated as the arithmetic mean of all task scores. The Gemma2-2B model failed to produce valid outputs during inference and was therefore excluded from the results.}
\label{tab:model_mt_comparison}
\begin{tabular}{lcccccccc}
\toprule
 & Llama3.3 & Phi4 & Qwen2.5 & Deepseek & Gemma2 & Mistral & Llama3.1 & Llama3.2 \\
\midrule
T01 & \textbf{0.971} & 0.960 & \textbf{0.971} & 0.939 & 0.964 & 0.924 & 0.905 & 0.500 \\
T02 & 0.788 & \textbf{0.834} & 0.811 & 0.792 & 0.812 & 0.808 & 0.564 & 0.500 \\
T03 & 0.923 & 0.905 & 0.902 & 0.888 & \textbf{0.945} & 0.841 & 0.570 & 0.500 \\
T04 & 0.500 & 0.566 & 0.515 & \textbf{0.686} & 0.509 & 0.538 & 0.495 & 0.500 \\
T05 & 0.840 & 0.850 & 0.829 & 0.804 & \textbf{0.854} & 0.818 & 0.584 & 0.500 \\
T06 & 0.708 & 0.655 & \textbf{0.730} & 0.699 & 0.594 & 0.663 & 0.541 & 0.500 \\
T07 & \textbf{0.955} & 0.838 & 0.892 & 0.895 & 0.765 & 0.850 & 0.673 & 0.500 \\
T08 & 0.883 & 0.818 & \textbf{0.890} & 0.847 & 0.735 & 0.743 & 0.608 & 0.500 \\
T09 & 0.619 & 0.701 & 0.707 & \textbf{0.730} & 0.606 & 0.472 & 0.415 & 0.141 \\
T10 & \textbf{0.978} & 0.948 & 0.955 & 0.957 & 0.814 & 0.777 & 0.754 & 0.144 \\
T11 & \textbf{0.669} & 0.566 & 0.620 & 0.526 & 0.370 & 0.490 & 0.366 & 0.236 \\
T12 & \textbf{0.842} & 0.838 & 0.835 & 0.834 & 0.788 & 0.780 & 0.796 & 0.046 \\
T13 & \textbf{0.736} & 0.697 & 0.710 & 0.685 & 0.646 & 0.700 & 0.535 & 0.233 \\
T14 & 0.573 & 0.647 & \textbf{0.702} & 0.511 & 0.511 & 0.381 & 0.423 & 0.017 \\
T15 & 0.917 & 0.906 & \textbf{0.931} & 0.890 & 0.881 & 0.868 & 0.861 & 0.505 \\
T16 & \textbf{0.959} & 0.937 & 0.893 & 0.920 & 0.879 & 0.821 & 0.596 & 0.503 \\
T17 & 0.767 & 0.740 & 0.733 & 0.753 & \textbf{0.768} & 0.674 & 0.608 & 0.131 \\
T18 & \textbf{0.732} & 0.649 & 0.540 & 0.638 & 0.497 & 0.515 & 0.215 & 0.030 \\
T19 & 0.995 & \textbf{0.997} & 0.995 & 0.996 & 0.912 & 0.996 & 0.967 & 0.065 \\
T20 & 0.991 & 0.986 & \textbf{0.993} & 0.988 & 0.812 & 0.966 & 0.861 & 0.059 \\
T21 & 0.993 & 0.996 & \textbf{0.998} & 0.997 & 0.958 & 0.996 & 0.986 & 0.269 \\
T22 & \textbf{0.976} & 0.960 & 0.968 & 0.966 & 0.917 & 0.948 & 0.892 & 0.126 \\
T23 & 0.955 & 0.889 & 0.953 & \textbf{0.974} & 0.813 & 0.952 & 0.800 & 0.308 \\
T24 & \textbf{0.961} & 0.958 & 0.872 & 0.936 & 0.912 & 0.927 & 0.716 & 0.278 \\
T25 & 0.028 & 0.169 & 0.095 & \textbf{0.173} & 0.044 & 0.010 & 0.003 & 0.000 \\
T26 & 0.401 & 0.407 & 0.305 & 0.253 & \textbf{0.411} & 0.243 & 0.262 & 0.221 \\
T27 & \textbf{0.467} & 0.439 & 0.426 & 0.428 & 0.441 & 0.419 & 0.279 & 0.247 \\
T28 & 0.161 & 0.162 & 0.178 & 0.122 & 0.104 & 0.139 & \textbf{0.195} & 0.031 \\
\midrule
$S_\mathrm{DRAGON}$ & \textbf{0.760} & 0.751 & 0.748 & 0.744 & 0.688 & 0.688 & 0.588 & 0.271 \\
\bottomrule
\end{tabular}
\end{table}

\clearpage

\begin{table}[h]
\centering
\caption{Model performance comparison on benchmark tasks, with and without machine translation (MT). Best score per model per task is bolded.}
\label{tab:model_performance_translation_abbrev}
\begin{tabular}{lcc|cc|cc}
\toprule
 &
\multicolumn{2}{c|}{\textbf{Phi4}} &
\multicolumn{2}{c|}{\textbf{Mistral}} &
\multicolumn{2}{c}{\textbf{Llama3.1}} \\
\textbf{Task} & w/o MT & w/ MT & w/o MT & w/ MT & w/o MT & w/ MT \\
\midrule
T01 & \textbf{0.960} & 0.893 & \textbf{0.905} & 0.890 & \textbf{0.890} & 0.566 \\
T02 & \textbf{0.834} & 0.686 & \textbf{0.808} & 0.766 & \textbf{0.564} & 0.476 \\
T03 & \textbf{0.905} & 0.740 & \textbf{0.841} & 0.825 & \textbf{0.570} & 0.489 \\
T04 & \textbf{0.566} & 0.558 & 0.538 & \textbf{0.627} & \textbf{0.495} & 0.473 \\
T05 & \textbf{0.850} & 0.712 & \textbf{0.818} & 0.784 & 0.584 & \textbf{0.697} \\
T06 & \textbf{0.655} & 0.638 & \textbf{0.663} & 0.581 & 0.541 & \textbf{0.567} \\
T07 & \textbf{0.838} & 0.612 & \textbf{0.850} & 0.758 & \textbf{0.673} & 0.526 \\
T08 & \textbf{0.818} & 0.722 & 0.743 & \textbf{0.768} & \textbf{0.608} & 0.562 \\
T09 & \textbf{0.701} & 0.559 & \textbf{0.472} & 0.420 & \textbf{0.415} & 0.193 \\
T10 & \textbf{0.948} & 0.853 & \textbf{0.777} & 0.538 & \textbf{0.754} & 0.051 \\
T11 & \textbf{0.566} & 0.466 & \textbf{0.490} & 0.285 & \textbf{0.366} & 0.053 \\
T12 & \textbf{0.838} & 0.691 & \textbf{0.780} & 0.333 & \textbf{0.796} & 0.130 \\
T13 & \textbf{0.697} & 0.471 & \textbf{0.700} & 0.270 & \textbf{0.535} & -0.007 \\
T14 & \textbf{0.647} & 0.133 & \textbf{0.381} & 0.073 & \textbf{0.423} & 0.009 \\
T15 & \textbf{0.906} & 0.884 & \textbf{0.868} & 0.831 & \textbf{0.861} & 0.498 \\
T16 & \textbf{0.937} & 0.650 & \textbf{0.821} & 0.742 & \textbf{0.596} & 0.482 \\
T17 & \textbf{0.740} & 0.604 & \textbf{0.674} & 0.653 & \textbf{0.608} & 0.322 \\
T18 & \textbf{0.649} & 0.571 & 0.515 & \textbf{0.535} & \textbf{0.215} & 0.117 \\
T19 & \textbf{0.997} & 0.392 & \textbf{0.996} & 0.831 & \textbf{0.967} & 0.508 \\
T20 & \textbf{0.986} & 0.470 & \textbf{0.966} & 0.862 & \textbf{0.861} & 0.696 \\
T21 & \textbf{0.996} & 0.530 & \textbf{0.996} & 0.817 & \textbf{0.986} & 0.459 \\
T22 & \textbf{0.960} & 0.688 & \textbf{0.948} & 0.907 & \textbf{0.892} & 0.560 \\
T23 & \textbf{0.952} & 0.668 & \textbf{0.952} & 0.902 & \textbf{0.800} & 0.492 \\
T24 & \textbf{0.958} & 0.465 & \textbf{0.927} & 0.793 & \textbf{0.716} & 0.341 \\
T25 & \textbf{0.169} & 0.008 & \textbf{0.010} & 0.009 & \textbf{0.003} & 0.003 \\
T26 & \textbf{0.407} & 0.065 & \textbf{0.243} & 0.064 & \textbf{0.262} & 0.038 \\
T27 & \textbf{0.439} & 0.179 & \textbf{0.419} & 0.191 & \textbf{0.279} & 0.130 \\
T28 & \textbf{0.162} & 0.006 & \textbf{0.195} & 0.002 & \textbf{0.002} & 0.000 \\ \hline
S\_DRAGON & \textbf{0.751} & 0.533 & \textbf{0.688} & 0.573 & \textbf{0.588} & 0.337 \\
\bottomrule
\end{tabular}
\end{table}

\clearpage

\begin{table}[h]
\centering
\caption{Model name abbreviations used in performance tables.}
\label{tab:model_legend}
\begin{tabular}{ll}
\toprule
Abbreviation & Full Model Name \\
\midrule
Llama3.3 & llama3.3:70b-instruct-q4\_K\_M\\
Phi4 & phi4:14b-q4\_K\_M \\
Qwen2.5 & qwen2.5:14b-instruct-q4\_K\_M \\
Deepseek & deepseek-r1:14b-qwen-distill-q4\_K\_M \\
Gemma2 & gemma2:9b-instruct-Q4\_0 \\
Mistral & mistral-nemo:12b-instruct-2407-q4\_0 \\
Llama3.1 & llama3.1:8b-instruct-q4\_K\_M \\
Llama3.2 & llama3.2:3b-instruct-q4\_K\_M \\
\bottomrule
\end{tabular}
\end{table}

\clearpage

\section*{Supplementary Note 4: Task-Specific Prompts}
\label{supp4}
This section outlines the task-specific prompts used to run models across the 28 tasks of the DRAGON challenge. These prompts were developed based on the task definitions provided in the supplementary materials of Bosma et al. (2025) \cite{bosma2025dragon}. Each prompt was provided as the description entry in the Taskfile of the llm\_extractinator framework. The framework supplements these prompts with an additional system message specifying the task name, task description, and expected output format. To ensure consistency and comparability across experiments, all models were evaluated using the same set of Taskfiles.

\subsection*{System Prompt}
\begin{lstlisting}[language=Python]
As an expert medical professional, your objective is to accurately evaluate the provided medical report and determine the following:

**Task:** \{task\}

**Description:** \{description\}

Please carefully review the report and think step by step.

It is essential to provide a confident and definitive answer. Avoid expressing uncertainty and make the most informed judgment based on the information presented.

\{output\_format\}
\end{lstlisting}

\subsection*{Task001: Adhesion Clf}

\begin{lstlisting}[language=Python]
This task involves analyzing radiology reports to identify whether presence of adhesions are mentioned. The output should be a binary classification: 'True' if adhesions are described, and 'False' if they are not.
\end{lstlisting}

\subsection*{Task002: Nodule Clf}

\begin{lstlisting}[language=Python]
This task requires analyzing the text of radiology reports to identify whether a pulmonary nodule is specifically mentioned. It is only relevant whether one is written literally in the text or not. You should not make inferences of the patient's health based on the report. The result should be a binary classification: 'True' if a nodule is mentioned, and 'False' if it is not.
\end{lstlisting}

\subsection*{Task003: Kidney Clf}

\begin{lstlisting}[language=Python]
This task involves determining whether a radiology report mentions any abnormalities related to the kidneys. Abnormalities include renal cell carcinoma, angiomyolipoma, cysts, kidney stones, conjoined kidneys, cases with partial or full nephrectomy, and several other rare abnormalities. The output should be a binary classification: 'True' if a kidney abnormality is mentioned, and 'False' if it is not.
\end{lstlisting}

\subsection*{Task004: Skin Case Selection Clf}

\begin{lstlisting}[language=Python]
This task requires evaluating radiology reports to determine if they meet exclusion criteria, such as the report being incomplete or containing an incomplete diagnosis. The output should be a binary classification: 'True' if the report matches exclusion criteria, and 'False' if it does not.
\end{lstlisting}

\subsection*{Task005: Recist Timeline Clf}

\begin{lstlisting}[language=Python]
This task involves analyzing radiology reports to determine whether the scan is a baseline (initial) scan or a follow-up scan. The result should be a binary classification: 'True' for a baseline scan and 'False' for a follow-up scan.
\end{lstlisting}

\subsection*{Task006: Pathology Tumor Origin Clf}

\begin{lstlisting}[language=Python]
This task involves analyzing pathology reports to determine if the cancer originated in the lung or if it is a metastasis from another organ. The output should be a binary classification: 'True' if the tumor originated in the lung, and 'False' if it did not.    
\end{lstlisting}

\subsection*{Task007: Nodule Diameter Presence Clf}

\begin{lstlisting}[language=Python]
This task involves analyzing radiology reports to check whether the pulmonary nodule mentioned in the report includes a size measurement. The result should be a binary classification: 'True' if a size is provided, and 'False' if it is not.
\end{lstlisting}

\subsection*{Task008: Pdac Size Presence Clf}

\begin{lstlisting}[language=Python]
This task involves analyzing radiology reports to check if a pancreatic ductal adenocarcinoma (PDAC) mentioned in the report includes a size measurement. The result should be a binary classification: 'True' if a size is provided, and 'False' if it is not.       
\end{lstlisting}

\subsection*{Task009: Pdac Diagnosis Clf}

\begin{lstlisting}[language=Python]
This task involves classifying the diagnosis mentioned in the radiology report into one of three categories: pancreatic ductal adenocarcinoma (PDAC), other pancreatic disease, or a normal pancreas.
\end{lstlisting}

\subsection*{Task010: Prostate Radiology Clf}

\begin{lstlisting}[language=Python]
This task involves analyzing the prostate radiology report to count the number suspicious lesions. Lesions are suspicious if they have a PI-RADS score of 3,4 or 5 lesions. The output should be the number of suspicious lesions, ranging from 0 to 4.
\end{lstlisting}

\subsection*{Task011: Prostate Pathology Clf}

\begin{lstlisting}[language=Python]
This task involves analyzing the prostate pathology report to count the number of lesions that have a Gleason score greater than or equal to 7. The output should be the number of such lesions, ranging from 0 to 3.
\end{lstlisting}

\subsection*{Task012: Pathology Tissue Type Clf}

\begin{lstlisting}[language=Python]
This task involves analyzing pathology reports to classify the type of tissue described. The output should be a classification into one of three categories: Biopsy, Resection, or Excision.
\end{lstlisting}

\subsection*{Task013: Pathology Tissue Origin Clf}

\begin{lstlisting}[language=Python]
This task involves extracting the origin of the material described in the pathology report. The output should classify the tissue origin into one of the following categories: lung, lymph node, bronchus, liver, brain, bone, or other. The origin of the tissue is generally mentioned at the beginning of the report as aard materiaal.
\end{lstlisting}

\subsection*{Task014: Textual Entailment Clf}

\begin{lstlisting}[language=Python]
This task involves analyzing pairs of sentences to determine their relationship. The output should classify the relationship as either contradiction, neutral, or entailment.
\end{lstlisting}

\subsection*{Task015: Colon Pathology Clf}

\begin{lstlisting}[language=Python]
For the given numeral, predict whether the specimen was obtained from 1) biopsy (true) or polypectomy (false), and whether the pathologist rated the specimen as 2) hyperplastic polyps, 3) low-grade dysplasia, 4) high-grade dysplasia, 5) cancer, 6) serrated polyps, or 7) non-informative. Give a true or false answer for each of the categories.
\end{lstlisting}

\subsection*{Task016: Recist Lesion Size Presence Clf}

\begin{lstlisting}[language=Python]
This task involves analyzing radiology reports to determine whether the size is mentioned for each of the 5 RECIST target lesions. The output should be a binary classification for each lesion: 'True' if the size is mentioned, and 'False' if it is not.
\end{lstlisting}

\subsection*{Task017: Pdac Attributes Clf}

\begin{lstlisting}[language=Python]
This task involves classifying the attributes of pancreatic ductal adenocarcinoma (PDAC) as described in the radiology report. The attributes to be classified include attenuation (iso/hyper/hypo/not mentioned) and location (head/body/tail/not mentioned). The output should provide a classification for both of these attributes.
\end{lstlisting}

\subsection*{Task018: Osteoarthritis Clf}

\begin{lstlisting}[language=Python]
This task involves classifying the Kellgren-Lawrence grade of osteoarthritis for both the left and right sides as described in the radiology report. The grades range from 0 to 4, with additional categories for 'not applicable (n)' and 'prosthesis (p)'. The output should provide a classification for each side.

Kellgren-Lawrence scale:

0: no radiographic core features of osteoarthritis, no joint gap narrowing, no bone abnormalities. Keywords: no coxarthrosis

1: possible joint gap narrowing, possible osteophyte formation. Keywords: no obvious coxarthrosis

2: obvious osteophyte formation, possible joint gap narrowing. Keywords: minimal coxarthrosis, incipient coxarthrosis, mild coxarthrosis, minor coxarthrosis

3: moderate osteophyte formation, marked joint gap narrowing and some sclerosis, possible degenerative bone defects. Keywords: moderate coxarthrosis

4: large definite osteophytes, definite joint gap narrowing and severe sclerosis, definite degenerative bone defects. Keywords: end-stage coxarthrosis, severe coxarthrosis, substantial coxarthrosis, strong coxarthrosis, obvious degeneration, obvious osteophyte formation

not applicable: there is not enough information in the report to give an assessment

prosthesis: the patient has a hip prosthesis.
\end{lstlisting}

\subsection*{Task019: Prostate Volume Reg}

\begin{lstlisting}[language=Python]
This task involves predicting the prostate volume in cubic centimeters, which is either directly described in the radiology report, or needs to be calculated based on the PSA density or the given measurements. All required information is provided in the report, and the PSA density is related to the PSA and prostate volume as: prostate volume = PSA / PSA density.
\end{lstlisting}

\subsection*{Task020: Psa Reg}

\begin{lstlisting}[language=Python]
This task involves estimating the Prostate-Specific Antigen (PSA) level based on the information provided in the radiology report. If a range is given, the average should be calculated.
\end{lstlisting}

\subsection*{Task021: Psad Reg}

\begin{lstlisting}[language=Python]
This task involves predicting the PSA density, which is either directly described in the radiology report or needs to be calculated based on the provided PSA level and prostate volume. The PSA density is related to the PSA and prostate volume as: PSA density = PSA / prostate volume.
\end{lstlisting}

\subsection*{Task022: Pdac Size Reg}

\begin{lstlisting}[language=Python]
This task involves estimating the size of pancreatic ductal adenocarcinoma (PDAC) as described in the radiology report, with the size given in millimeters.
\end{lstlisting}

\subsection*{Task023: Nodule Diameter Reg}

\begin{lstlisting}[language=Python]
This task involves estimating the diameter of the largest pulmonary nodule described in the radiology report, with the diameter given in millimeters. When multiple sizes are described for a single lesion (e.g., the short and long axis), the size for that lesion should be averaged (e.g., 9 mm for a lesion of size 1.0 x 0.8 cm).
\end{lstlisting}

\subsection*{Task024: Recist Lesion Size Reg}

\begin{lstlisting}[language=Python]
This task involves estimating the size of each of the up to 5 RECIST target lesions described in the radiology report, with the size given in millimeters. For lymph nodes, the short axis should be reported. If less than 5 lesions are described, the remaining lesion sizes should be set to 0.
\end{lstlisting}

\subsection*{Task025: Anonymisation Ner}

\begin{lstlisting}[language=Python]
Identify and classify sequences of tokens in the given text that qualify as Personally Identifiable Information (PII). Create a list of lists where each inner list contains two entries: 1. The exact sequence of text that qualifies as PII (e.g., '5 maart 2023', 'Jan Jansen', or 'RPT-12345'). 2. The corresponding predefined category tag (e.g., \texttt{<DATUM>}, \texttt{<PERSOON>}, \texttt{<RAPPORT\_ID>}, etc.). If no PII entities are present in the text, return an empty list. The model should be accurate in its identification and classification, ensuring entities are tagged correctly based on the predefined categories. Predefined PII Categories: 1. Dates (\texttt{<DATUM>}): Includes specific calendar dates. 2. Person Names (\texttt{<PERSOON>}): Full names or identifiable portions of names. 3. Report Identifiers (\texttt{<RAPPORT\_ID>}): Alphanumeric or symbolic identifiers assigned to reports. 4. Places (\texttt{<PLAATS>}): Names of locations such as cities, countries, addresses, or landmarks. 5. Personally Identifying Numbers (\texttt{<PHINUMMER>}): Numbers uniquely tied to an individual, including Social Security Numbers (SSNs), Tax Identification Numbers (TINs), passport numbers, or other similar identifiers. 6. Clinical Trial Names (\texttt{<STUDIE\_NAAM>}): Official names of clinical trials or studies. 7. Hospital Accreditation Numbers (\texttt{<ACCREDITATIE\_NUMMER>}): Unique codes or numbers issued to hospitals or healthcare institutions as part of accreditation. 8. Times (\texttt{<TIJD>}): Specific times of day, including those with time zones. 9. Patient Ages (\texttt{<LEEFTIJD>}): Exact ages or references to ages that directly identify an individual. Instructions: Identify sequences of text that represent PII, append a list containing the text and its corresponding category tag to the output list, and return the list. If no PII entities are detected, return an empty list ([]). Ensure each entity is tagged correctly, avoid false positives, and do not infer entities beyond what is explicitly stated.
\end{lstlisting}

\subsection*{Task026: Medical Terminology Ner}

\begin{lstlisting}[language=Python]
Your task is to identify sequences of tokens in the given text that represent medical terminology. For each identified term, provide its exact text as it appears in the input. The output should be a list of medical terminology entities in the form of a single list of strings, where each string represents one identified medical term. Ensure accuracy by only identifying terms that are clearly medical in nature, avoiding any ambiguity or overlap with non-medical language. By adhering to these instructions, you will deliver a structured and accurate identification of medical terminology entities found in the text.
\end{lstlisting}

\subsection*{Task027: Prostate Biopsy Ner}

\begin{lstlisting}[language=Python]
Your task is to analyze prostate biopsy reports to identify and classify sequences of words that describe the location of each numbered biopsy and to determine whether the lesion sampling was REPRESENTATIVE (representatief), NOT REPRESENTATIVE (niet representatief), or AMBIGUOUS (ambigu). The output should be a list of each biopsy, where for each biopsy, you include: 1) the biopsy number as an integer (converted from Roman numerals I, II, III, etc.), 2) the exact words that describe the biopsy location, 3) the quality of the biopsy sampling (representatief, niet representatief, ambigu), and 4) the exact words that describe the quality. Ensure all classifications are accurate and based solely on the information in the text. Example Output Format: [\{\{"number": 1, "location": "left apex", "quality": "REPRESENTATIVE", "quality\_description": "adequate tissue sample"\}\}, \{\{"number": 2, "location": "right base", "quality": "NOT REPRESENTATIVE", "quality\_description": "insufficient tissue"\}\}]. By adhering to these instructions, you will deliver a structured and detailed analysis of the biopsy report.
\end{lstlisting}

\subsection*{Task028: Skin Pathology Ner}

\begin{lstlisting}[language=Python]
Your task is to analyze each word in a skin pathology report to classify and split the diagnosis for each specified case, numbered from 1 to 20. For each case, you should identify: 1) the case number as an integer, 2) the diagnosis, which can be "BCC", "Benign", or "Other", including the exact words from the text describing the diagnosis, 3) any subtypes present for cases diagnosed with basal cell carcinoma, including the exact words from the text describing the subtypes, and 4) the tissue acquisition method (either "biopt" or "excision"), including the exact words from the text describing the method. The output should be a list of dictionaries, with each dictionary containing the details for one case. Ensure all classifications and text references are accurate and derived directly from the input. By adhering to these instructions, you will deliver a structured and detailed analysis of each case in the pathology report, ensuring the exact words from the text are captured for each category.
\end{lstlisting}

\end{document}